\documentclass[conference]{IEEEtran}
\IEEEoverridecommandlockouts

\usepackage{cite}
\usepackage{amsmath,amssymb,amsfonts}
\usepackage{algorithmic}
\usepackage{graphicx}
\usepackage{textcomp}
\usepackage{xcolor}
\def\BibTeX{{\rm B\kern-.05em{\sc i\kern-.025em b}\kern-.08em
    T\kern-.1667em\lower.7ex\hbox{E}\kern-.125emX}}

\usepackage[pagebackref=false,breaklinks=true,colorlinks,citecolor=blue,bookmarks=false]{hyperref}
\usepackage{caption}
\usepackage{booktabs}
\usepackage{pifont}
\usepackage{bm}
\usepackage{multirow}
\usepackage{graphicx}
\usepackage{indentfirst}
\usepackage{lipsum} 
\usepackage{array}    
\usepackage{subcaption} 
\usepackage{makecell}

 \usepackage{bbding}

\title{Event-based Video Person Re-identification via Cross-Modality and Temporal Collaboration}
\author{Renkai Li$^{1}$ \hspace{0.1in} Xin Yuan$^{1,2,*}$\thanks{*Corresponding author (yuanxincherry@gmail.com; liuwei@wust.edu.cn).} \hspace{0.1in} Wei Liu$^{1,*}$  \hspace{0.1in} Xin Xu$^{1,2}$\hspace{0.1in}\\
$^{1}$ School of Computer Science and Technology, Wuhan University of Science and Technology\\
$^{2}$ Hubei Province Key Laboratory of Intelligent Information Processing and Real-Time Industrial System

}

\begin{document}
%
\maketitle
\begin{abstract}
Video-based person re-identification (ReID) has become increasingly important due to its applications in video surveillance applications. By employing events in video-based person ReID, more motion information can be provided between continuous frames to improve recognition accuracy. Previous approaches have assisted by introducing event data into the video person ReID task, but they still cannot avoid the privacy leakage problem caused by RGB images. In order to avoid privacy attacks and to take advantage of the benefits of event data, we consider using only event data. To make full use of the information in the event stream, we propose a Cross-Modality and Temporal Collaboration (CMTC) network for event-based video person ReID. First, we design an event transform network to obtain corresponding auxiliary information from the input of raw events. Additionally, we propose a differential modality collaboration module to balance the roles of events and auxiliaries to achieve complementary effects. Furthermore, we introduce a temporal collaboration module to exploit motion information and appearance cues. Experimental results demonstrate that our method outperforms others in the task of event-based video person ReID.
\end{abstract}



\section{Introduction}
\label{sec:intro}
\noindent
Person re-identification (ReID) is a popular technique that has emerged in the field of computer vision in the last few years, which utilizes the techniques of computer vision to determine the presence of a specific person being tracked in a set of images or a video sequence. Image-based person ReID \cite{zhu2024generative,1,3,wang2020re} has made significant progress in recent years. With the popularity of video capture systems and the introduction of large-scale video benchmarks \cite{vid1,vid2}, the video-based person ReID has also gradually attracted wide attention. Compared with images, video data usually contains richer temporal and spatial information, which can be used to enhance the appearance of people \cite{5,6,7}, and thus improve the performance of the person ReID task.

Although video data can provide extremely rich information for person ReID task, it is impossible to avoid the effects of privacy information leakage, motion blur, object occlusion, and changes in illumination conditions, which can lead to the absence of key recognition cues~\cite{liu2024mix,gao2024part,yuan2023searching}. For the reasons stated above, we turn our attention to event cameras \cite{8,9,10,11}, as a new type of asynchronous image sensor, that has the advantages of high frame rate, high dynamic range, and low power consumption. Unlike traditional fixed frame rate cameras, event cameras record only the asynchronous changes in the captured scene thereby greatly reducing the sensor latency. Since event data discards redundant visual information, it can be more protective of the subject's privacy \cite{ahmad}, thus providing a certain degree of protection against the problem of easily leaking privacy in visual applications in public spaces.

 \begin{figure}
\centering
\includegraphics[width=1.0\linewidth]{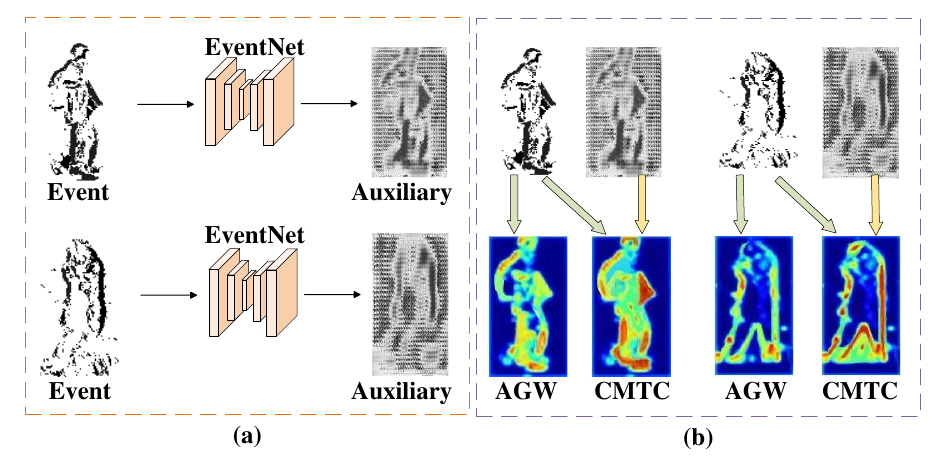}\
\caption{(a) shows the auxiliary information generated by the original event after passing through our EventNet. (b) shows two specific cases, in each case giving the original event, the auxiliary information corresponding to the event, the feature map of AGW, and the feature map of our CMTC.} %
\label{fig:1}
\end{figure}

Traditional approaches to processing event data have mainly involved processing event information through probabilistic filters as well as spiking neural networks (SNNs) \cite{e12,4}, or converting events into tensor structures compatible with past algorithms for visual tasks as input \cite{maqueda2018event}. Previously, Ahmad \textit{et al.} \cite{ahmad} proposed and validated the feasibility of event camera for image-based person ReID, and there is currently no method for video-based person ReID task based on event data, so we start this work based on this point. Fig.~\ref{fig:1} illustrates two examples of generated auxiliaries and gives feature maps of AGW \cite{e8} and CMTC.

The main contributions of this work are summarized as follows: (1) we design EventNet for generating Auxiliaries and apply the Auxiliaries along with Events as inputs to event-based video person ReID task, (2) we propose the Cross-Modality and Temporal Collaboration (CMTC) method to fully utilize the complementary nature of Events and Auxiliaries as well as the temporal information between consecutive frames, (3) we demonstrate our method's performance improvement on several public datasets.


\begin{figure*}[ht]
\centering
\includegraphics[width=0.9\linewidth]{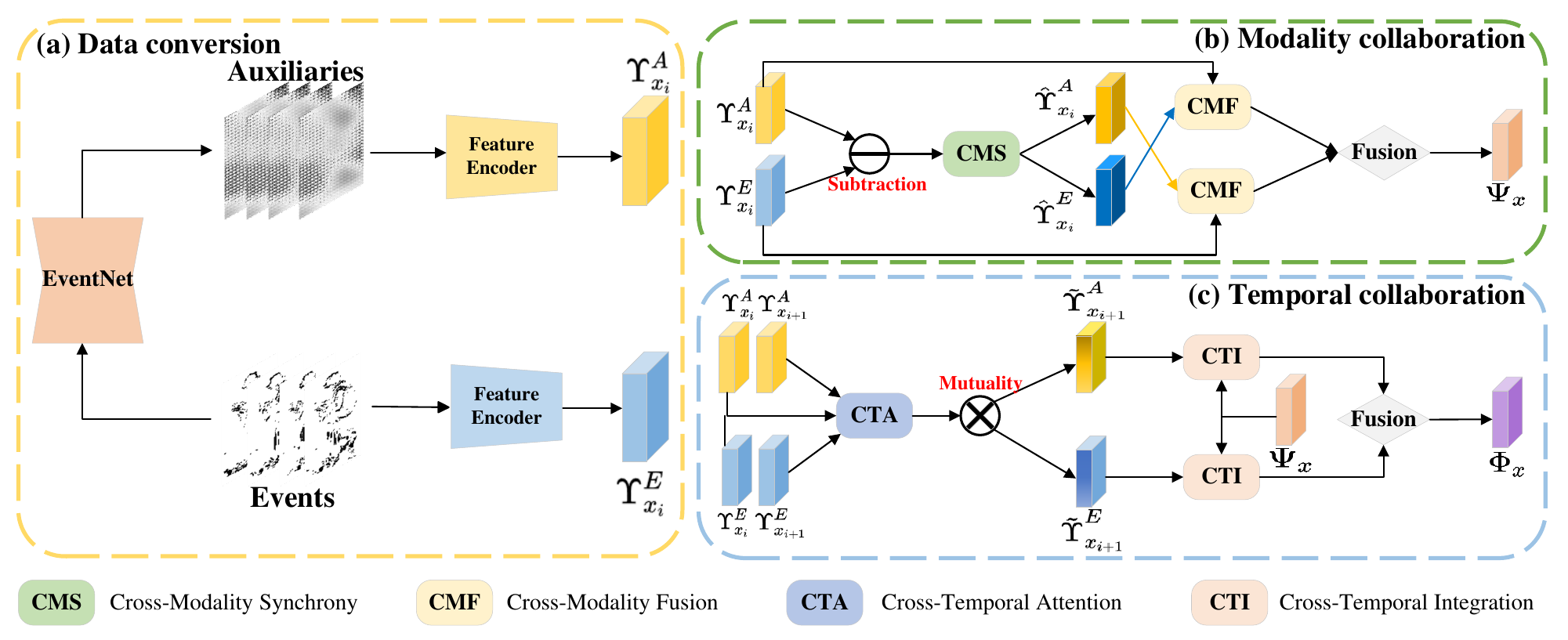}\
\caption{The overview of our Cross-Modality and Temporal Collaboration Network (CMTC).} %
\label{fig:2}
\end{figure*}


\section{Proposed Method}
\label{sec:method}

\subsection{Overview} 
\noindent
As shown in Fig.~\ref{fig:2}, our approach takes event data as input to EventNet, and the Auxiliaries generated by EventNet will be used as input to the subsequent modules together with the event data through the feature encoder to form event feature $\Upsilon_{x_i}^{E}$ and auxiliary feature $\Upsilon_{x_i}^{A}$. $\Upsilon_{x_i}^{E}$ and $\Upsilon_{x_i}^{A}$ enter the Modality Collaboration Module to facilitate the collaborative fusion of the two, which is complemented and enhanced by the difference operation and CMS, and then equalized by the CMF to generate $\Psi_{x}$. In addition, the neighboring events and the corresponding auxiliary information are input to the CTA, and then the output and $\Psi_{x}$ are input to the CTI and feature fusion is performed to obtain ${\Phi}_{x}$.


\subsection{EventNet} 
\noindent
EventNet is our deep-learning network designed for event-based video person ReID task, capable of processing event data and extracting useful feature information. The network structure consists of convolutional layers (Conv + LeakyReLU), average pooling layers, and bilinear upsampling layers, with the input data being our event dataset of images (Events) and the output data being auxiliary information that can be seen in the event contours (Auxiliaries). Dimensionality reduction and reconstruction operations are performed through average pooling and up-sampling operations to preserve the valid information of the data.

\subsection{Modality Collaboration} 
\noindent
Modality Collaboration (MC) mainly contains Cross-Modality Synchronization (CMS) and Cross-Modality Fusion (CMF). Cross-Modality Synchronization (CMS) is designed because there will be effective complementary information as well as redundant information in the data between different modalities. We perform the different operations between $\Upsilon_{x_i}^{E}$ and $\Upsilon_{x_i}^{A}$ to guide CMS, and use different modalities to obtain the attention map. Feature enhancement is achieved by multiplying the attention map with features.

First, the inputs $\Upsilon_{x_i}^{E}$ and $\Upsilon_{x_i}^{A}$ are taken to derive the differential modality D by subtracting them: 
\begin{equation}
D = \Upsilon_{x_i}^{E} - \Upsilon_{x_i}^{A}
\end{equation}

By introducing the attention mechanism \cite{attention}, $\Upsilon_{x_i}^{E}$ and $\Upsilon_{x_i}^{A}$ are converted into key $K_{E}$, value $V_{E}$ and key $K_{A}$, value $V_{A}$. D is converted into query $Q_{D}$. Based on these matrices obtained after the conversion to compute our desired attention graph and derive the augmented features $\hat{\Upsilon} _{x_i}^{E}$ and $\hat{\Upsilon} _{x_i}^{A}$ after CMS:
\begin{equation}
\begin{split}
\hat{\Upsilon} _{x_i}^{E} &= Softmax(Q_{D}K_{E}^{T})V_{E} \\
\hat{\Upsilon} _{x_i}^{A} &= Softmax(Q_{D}K_{A}^{T})V_{A}
\end{split}
\end{equation}

Cross-Modality Fusion (CMF) is designed to set the contribution weights of different modalities according to the importance of the information within the modality. Since our inputs are symmetric, only the case of taking $\hat{\Upsilon} _{x_i}^{E}$ and $\Upsilon_{x_i}^{A}$ as inputs is presented here. Firstly, $\hat{\Upsilon} _{x_i}^{E}$ and $\Upsilon_{x_i}^{A}$ are subjected to a series of operations such as convolution to obtain the weight parameter $W_{E}$ (similarly we can get $W_{A}$):
\begin{equation}
W_{E} = Avg(Sig(Conv(Concat(\hat{\Upsilon} _{x_i}^{E},\Upsilon_{x_i}^{A}))))
\end{equation}

The obtained weight parameters $W_{E}$ are multiplied with $\hat{\Upsilon} _{x_i}^{E}$ and $\Upsilon_{x_i}^{A}$ to get the features $\overline{\hat{\Upsilon} _{x_i}^{E}}$ and $\overline{\hat{\Upsilon} _{x_i}^{A}}$:
\begin{equation}
\overline{\hat{\Upsilon}_{x_i}^{E}}=\hat{\Upsilon} _{x_i}^{E}*W_{E} ,\quad  \overline{\hat{\Upsilon} _{x_i}^{A}}=\hat{\Upsilon}_{x_i}^{A}*W_{A}
\end{equation}

And then emphasize the features by using Channel and Spatial attention:
\begin{equation}
\overline{\hat{\Upsilon} _{x_i}^{E}}' = Channel(\overline{\hat{\Upsilon} _{x_i}^{E}}) * \overline{\hat{\Upsilon} _{x_i}^{E}}, \quad
\overline{\hat{\Upsilon} _{x_i}^{E}}'' = Spatial(\overline{\hat{\Upsilon} _{x_i}^{E}}') * \overline{\hat{\Upsilon} _{x_i}^{E}}'
\end{equation}
then we get the feature $F_{\alpha}$ = $\overline{\hat{\Upsilon} _{x_i}^{A}}$ + $\overline{\hat{\Upsilon} _{x_i}^{E}}''$(similarly we can get $F_{\beta}$).

Combining $F_{\alpha}$ and $F_{\beta}$ yields the final fusion feature $\Psi_{x}$:
\begin{equation}
\Psi_{x} = Concat(F_{\alpha},F_{\beta})
\end{equation}


\subsection{Temporal Collaboration} 
\noindent
Temporal Collaboration (TC) mainly contains Cross-Temporal Attention (CTA) and Cross-Temporal Integration (CTI).

Cross-Temporal Attention (CTA) is designed to better use the rich motion information between neighboring event frames. We take $\Upsilon_{x_i}^{E}$,$\Upsilon_{x_{i+1}}^{E}$ and $\Upsilon_{x_i}^{A}$,$\Upsilon_{x_{i+1}}^{A}$ of neighboring moments as inputs to the CTA. According to the same attention mechanism \cite{attention} as in CMS, $\Upsilon_{x_i}^{E}$ is converted to query $Q_{x_i}^{E}$,$\Upsilon_{x_{i+1}}^{E}$ to key $K_{x_{i+1}}^{E}$, value $V_{x_{i+1}}^{E}$, and $\Upsilon_{x_i}^{A}$ to query $Q_{x_i}^{A}$,$\Upsilon_{x_{i+1}}^{A}$ to key $K_{x_{i+1}}^{A}$, value $V_{x_{i+1}}^{A}$. Operate on these values to obtain augmented features $t_{1}$ and $t_{2}$:
\begin{equation}
\begin{split}
t_{1} = Softmax(Q_{x_i}^{E}(K_{x_{i+1}}^{E})^{T}) \\ 
t_{2} = Softmax(Q_{x_i}^{A}(K_{x_{i+1}}^{A})^{T}) \\
\end{split}
\end{equation}

And then, we use augmented features $t_{1}$ and $t_{2}$ to obtain the cross-temporal attention T:
\begin{equation}
T = t_{1} \cdot t_{2} 
\end{equation}

The features $\tilde{\Upsilon} _{x_{i+1}}^{E}$ and $\tilde{\Upsilon} _{x_{i+1}}^{A}$ are derived by multiplying $V_{x_{i+1}}^{E}$ and $V_{x_{i+1}}^{A}$ with  T:
\begin{equation}
\tilde{\Upsilon} _{x_{i+1}}^{E}=T*V_{x_{i+1}}^{E},\qquad   \tilde{\Upsilon} _{x_{i+1}}^{A}=T*V_{x_{i+1}}^{A}
\end{equation}

Cross-Temporal Integration (CTI) is designed to use the features $\Psi_x$ obtained from MC to enable the network to better learn the relationship between neighboring frames of different modalities. Since our input is symmetric, only one of these cases is presented. CTI obtains the feature representation P by averaging the fused features $\Psi_{x}$ that are obtained from MC and $\tilde{\Upsilon} _{x_{i+1}}^{E}$ by average pooling and performing splicing and convolution operations:
\begin{equation}
P=Conv(Concat(AvgPool(\Psi_{x}),AvgPool(\tilde{\Upsilon} _{x_{i+1}}^{E})))
\end{equation}

Multiplying $\Psi_{x}$ and $\tilde{\Upsilon} _{x_{i+1}}^{E}$ with P respectively and summing them gives us the feature $F_{\varphi}$ of the branch (similarly we can get the feature $F_{\eta}$):
\begin{equation}
F_{\varphi}=\Psi_{x}P+\tilde{\Upsilon} _{x_{i+1}}^{E}P
\end{equation}

Combining $F_{\varphi}$ and $F_{\eta}$ yields the final fusion feature $\Phi_{x}$:
\begin{equation}
\Phi_{x} = Concat(F_{\varphi},F_{\eta})
\end{equation}


\section{Experiments}
\label{sec:experiment}
\noindent
In this section, we describe the datasets used for the experiments and evaluate the performance of multiple methods from previous years on these datasets. We will provide details of the experiment and perform ablation experiments to demonstrate the effectiveness of our method.

\subsection{Datasets}   
\noindent
We use Event Camera Emulator V2E \cite{v2e} to evaluate performance by converting three video-based person ReID public datasets to event style, including PRID-2011 \cite{prid}, iLIDS-VID \cite{ilids-vid}, and MARS \cite{mars}.


\subsection{Evaluation Metrics}                   
\noindent
We use two measurement metrics to assess the accuracy of event-based video person ReID on given datasets: cumulative matching characteristics (CMC) based on Rank-1 and mean average precision (mAP).

\begin{table*}[ht]
\centering
\caption{Performance comparison of different methods on our event datasets: PRID-2011, iLIDS-VID and MARS. Best performance results are marked in \textbf{bold}.}
\label{tab:performance}
\resizebox{\linewidth}{!}{
\begin{tabular}{ l|c|cccc|cccc|cccc } 
\toprule
\multirow{2}{*} {Method} & \multirow{2}{*} {Venue} & \multicolumn{4}{c}{PRID-2011} & \multicolumn{4}{c}{ILIDS-VID} & \multicolumn{4}{c}{MARS} \\
\cmidrule(lr){3-6} \cmidrule(lr){7-10} \cmidrule(lr){11-14}
& & Rank1 & Rank5 & Rank10 & mAP & Rank1 & Rank5 & Rank10 & mAP & Rank1 & Rank5 & Rank10 & mAP\\
\midrule
OSNet\cite{e1} & ICCV 19 & 10.1 & - & - & 22.2 & 16.7 & - & - & 27.9 & 19.3 & - & - & 30.9 \\
SRS-Net\cite{e2} & TNNLS 20 & 9.0 & - & - & 17.2 & 19.3 & - & - & 32.7 & 10.0 & - & - & 20.9 \\
TCLNet\cite{e3} & ECCV 20 & 52.8 & 79.7 & 89.8 & 64.4 & 31.3 & 58.0 & 66.6 & 43.8 & 25.3 & 51.3 & 65.3 & 38.2 \\
CTL\cite{e4} & CVPR 21 & 13.5 & - & - & 20.4 & 18.0 & - & - & 28.4 & 12.7 & - & - & 25.6 \\
STMN\cite{e5} & ICCV 21 & 11.2 & - & - & 20.2 & 12.7 & - & - & 23.5 & 10.0 & - & - & 22.4 \\
GRL\cite{e6} & CVPR 21 & 11.2 & - & - & 21.4 & 18.0 & - & - & 30.2 & 16.7 & - & - & 27.7 \\
PSTA\cite{e7} & ICCV 21 & 30.3 & 61.8 & 75.3 & 46.4 & 14.7 & 28.7 & 47.3 & 23.7 & 12.0 & 24.1 & 43.3 & 22.7 \\
AGW\cite{e8} & TPAMI 21 & 57.9 & 84.3 & 91.7 & 69.7 & 34.1 & 62.0 & 71.3 & 45.8 & 30.6 & 57.6 & 67.2 & 42.5 \\
BiCNet-TKS\cite{e9} & CVPR 21 & 38.2 & 67.4 & 82.0 & 52.1 & 19.3 & 38.6 & 57.3 & 30.8 & 17.3 & 45.3 & 60.0 & 30.9 \\
MFA\cite{e10} & TIP 22 & 40.9 & 68.3 & 83.1 & 54.2 & 21.4 & 40.3 & 57.7 & 31.7 & 16.9 & 44.5 & 58.7 & 30.3 \\
FastRelD\cite{e11} & ACMMM 23 & 54.1 & 79.4 & 89.5 & 65.8 & 29.6 & 53.2 & 64.1 & 41.8 & 26.2 & 49.7 & 57.3 & 39.3 \\
SDCL\cite{e12} & CVPR 23 & 50.7 & 75.7 & 86.8 & 61.1 & 30.7 & 57.3 & 68.1 & 43.1 & 26.8 & 52.5 & 63.1 & 38.3 \\
VSLA-CLIP\cite{e13} & ECCV 24 & 45.3 & 70.7 & 84.2 & 57.5 & 26.6 & 48.4 & 62.0 & 36.9 & 20.9 & 38.3 & 55.1 & 33.4 \\
TF-CLIP\cite{e14} & AAAI 24 & 52.4 & 77.7 & 89.1 & 64.7 & 32.6 & 61.4 & 69.0 & 44.2 & 29.7 & 55.0 & 66.8 & 41.9 \\
\midrule
CMTC (ours) & This work & \textbf{60.4} & \textbf{91.5} & \textbf{93.3} & \textbf{71.9} & \textbf{35.3} & \textbf{64.3} & \textbf{74.0} & \textbf{47.5} & \textbf{31.6} & \textbf{59.5} & \textbf{69.4} & \textbf{43.8}\\
\bottomrule
\end{tabular}}    
\end{table*}

\subsection{Implementation Details} 
\noindent
EventNet's loss function uses MSELoss and a perceptual loss. The perceptual loss uses a pre-trained VGG network to extract features and compare the feature representations of the input and output. Our training strategy is to extract 8 frames from each video sequence and use them as input to EventNet. Events and the output Auxiliaries are resized to 256 $\times$ 128, and then used as subsequent inputs. We set the batch size to 32 and the initial learning rate to 0.0003, and trained the model by gradually decreasing the learning rate with a strategy of decaying 0.1 every 50 epochs. We use the Adam optimizer for 400 epochs to ensure that the model converges efficiently during the learning process.


\subsection{Comparison with Other Methods} 
\begin{figure}
\centering
\includegraphics[width=1.0\linewidth]{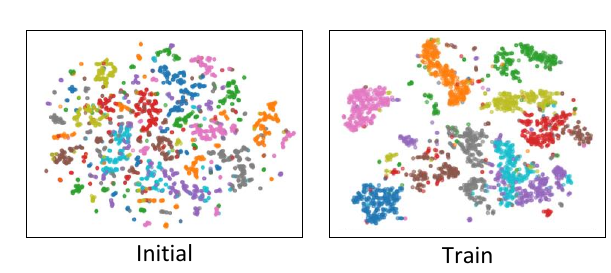}
\caption{The t-SNE visualization results for ten randomly selected identities on the PRID-2011 dataset, with different colors representing different identities.} %
\label{fig:3}
\end{figure}

\begin{figure}
\centering
\includegraphics[width=1.0\linewidth]{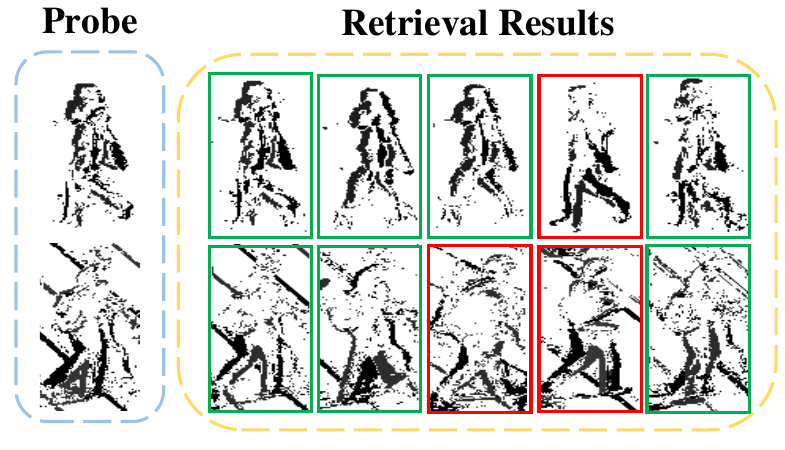}
\caption{Ranking results of CMTC. The red and green bounding indicate the error and correct result, respectively.} %
\label{fig:4}
\end{figure}

\noindent
In this study, three datasets are used to evaluate the effectiveness of our method. Table \ref{tab:performance} shows the results of the comparison of CMTC and other methods on event datasets PRID-2011\cite{prid}, iLIDS-VID\cite{ilids-vid} and MARS\cite{mars}. For PRID-2011, the smaller differences in perspective as well as less redundant information in the events resulted in a greater enhancement compared to the other two datasets. For iLIDS-VID and MARS, too much background noise in the events not only leads to an increase in the complexity of events recognition, but also interferes with the process of generating auxiliaries, resulting in a limited improvement in Rank-1 and mAP metrics. We have experimented by adopting an end-to-end approach, and achieved the best performance after utilizing the auxiliaries for feature-level fusion enhancement through the Modality Collaboration and Temporal Collaboration modules. We achieved 2.5$\%$, 1.2$\%$ and 1.0$\%$ Rank-1 improvement and 2.2$\%$, 1.7$\%$ and 1.3$\%$ mAP improvement on PRID-2011 \cite{prid}, iLIDS-VID \cite{ilids-vid} and MARS \cite{mars}. Fig.~\ref{fig:3} and Fig.~\ref{fig:4} show the t-SNE visualization result and the ranking result of our approach. These results of the above experiments demonstrate the effectiveness of our approach to the validity of auxiliaries and the usefulness of our approach on the task of event-based video person ReID.


\subsection{Ablation Study} 
\noindent
To evaluate the validity of the designed key components (EventNet, MC and TC) in the network, we performed ablation studies on PRID-2011~\cite{prid} and iLIDS-VID~\cite{ilids-vid} datasets. Table \ref{tab:ablation} shows the ablation results. First, we set a baseline for the experiment by simply inputting via event data, ignoring auxiliaries and temporal correlations between inputs. The first row shows the performance of the baseline, and the second row shows that our EventNet is effective. We attach the MC module to the baseline and perform cross-modality fusion of Events with the Auxiliaries. A comparison based on the second row and the third row shows that the MC module is able to effectively fuse the Events with Auxiliaries. Attaching the proposed TC module to the baseline enables the network to obtain time-related information from neighboring frames, thus improving the accuracy of our task. Comparing the second row and the fourth row can demonstrate the effectiveness of our proposed TC module.


\begin{table}[t]
\centering
\caption{Ablation studies on our event datasets: PRID-2011 and iLIDS-VID. Best results are marked in \textbf{bold}.}
\label{tab:ablation} 
\vspace{10pt}
{\fontsize{9pt}{12pt}\selectfont
\begin{tabular}{ l|cc|cc }
\toprule
\multirow{2}{*} { EventNet } \quad \multirow{2}{*} { MC } \quad \multirow{2}{*} {TC} & \multicolumn{2}{c}{PRID-2011} & \multicolumn{2}{c}{ILIDS-VID} \\
\cmidrule(lr){2-3} \cmidrule(lr){4-5}
& Rank1 & mAP & Rank1 & mAP \\
\midrule
\qquad \XSolidBrush \quad \qquad \XSolidBrush \qquad \XSolidBrush& 57.9 & 69.7 & 34.1 & 45.8 \\
\qquad \Checkmark \qquad \quad \XSolidBrush \qquad \XSolidBrush& 58.2 & 70.1 & 34.3 & 46.3 \\
\qquad \Checkmark \qquad \quad \Checkmark \qquad \XSolidBrush& 59.6 & 71.3 & 34.9 & 47.0 \\
\qquad \Checkmark \qquad \quad \XSolidBrush \qquad \Checkmark& 58.8 & 70.5 & 34.4 & 46.3 \\
\qquad \Checkmark \qquad \quad \Checkmark \qquad \Checkmark& \textbf{60.4} & \textbf{71.9} & \textbf{35.3} & \textbf{47.5} \\
\bottomrule
\end{tabular}
}
\vspace{-1mm}
\end{table}


\section{Conclusion}
\label{sec:Conclusion}
\noindent
In this work, we propose a Cross-Modality and Temporal Collaboration (CMTC) method that learns complementary modality features and temporal correlations from events and auxiliaries to handle the event-based video person ReID task. We design EventNet to generate auxiliaries from events. In contrast to past event-based ReID models, our approach deploys auxiliaries into dense event features to leverage the additional complementary information to guide video person ReID. To help the model discover more effective temporal features, we design a temporal collaboration module to preserve motion information between neighboring events. Experimental results demonstrate the effectiveness of our CMTC method for the event-based video person ReID task.


\small
\noindent \textbf{Acknowledgments:}
This work was supported by the National Nature Science Foundation of China (No. 62376201). This research was financially supported by funds from Hubei Province Key Laboratory of Intelligent Information Processing and Real-time Industrial System (Wuhan University of Science and Technology) (No. ZNXX2023QNO3), Fund of Hubei Key Laboratory of Inland Shipping Technology and Innovation (NO. NHHY2023004), Key Laboratory of Social Computing and Cognitive Intelligence (Dalian University of Technology), Ministry of Education (No. SCCI2024YB02), and Entrepreneurship Fund for Graduate Students of Wuhan University of Science and Technology (No. JCX2022031, JCX2023049, JCX2023160). We thank all reviewers for their comments.


\bibliographystyle{IEEEtran}
\bibliography{references}

\end{document}